\documentclass[journal]{IEEEtran}
\ifCLASSINFOpdf
\else
\fi
\usepackage{amssymb}
\usepackage{graphicx}
\usepackage{mathtools}
\usepackage{amsmath}
\usepackage{gensymb}
\usepackage{float}
\usepackage{multirow}
\usepackage{hyperref}
\usepackage{makecell}
\usepackage[utf8]{inputenc}
\usepackage[table]{xcolor}
\usepackage[english]{babel}
\usepackage[linesnumbered,ruled,vlined]{algorithm2e}

\SetCommentSty{mycommfont}

\SetKwInput{KwInput}{Input}                % Set the Input
\SetKwInput{KwOutput}{Output}              % set the Output
\usepackage{amsthm}

\usepackage{subcaption}

\newcommand{\gray}[1]{\textcolor{gray}{#1}}

\newcommand{\taninv}{\tan^{-1}}
\newcommand{\no}{\noindent}

\newcommand{\mc}[1]{\mathcal{#1}}
\newcommand{\bb}[1]{\mathbb{#1}}

\begin{document}
\title{COVID-Robot: Monitoring Social Distancing Constraints in Crowded Scenarios}

\author{Adarsh Jagan Sathyamoorthy, Utsav Patel, Yash Ajay Savle$^{*}$, Moumita Paul$^{*}$ and Dinesh Manocha.% <-this % stops a space

\thanks{* Authors contributed equally.}% <-this % stops a space
% \thanks{J. Doe and J. Doe are with Anonymous University.}% <-this % stops a space
% \thanks{Manuscript received April 19, 2005; revised August 26, 2015.}
}

\maketitle

\begin{abstract}
Maintaining social distancing norms between humans has become an indispensable precaution to slow down the transmission of COVID-19. We present a novel method to automatically detect pairs of humans in a crowded scenario who are not adhering to the social distance constraint, i.e. about 6 feet of space between them. Our approach makes no assumption about the crowd density or pedestrian walking directions. We use a mobile robot with commodity sensors, namely an RGB-D camera and a 2-D lidar to perform collision-free navigation in a crowd and estimate the distance between all detected individuals in the camera's field of view. In addition, we also equip the robot with a thermal camera that wirelessly transmits thermal images to a security/healthcare personnel who monitors if any individual exhibits a higher than normal temperature. In indoor scenarios, our mobile robot can also be combined with static mounted CCTV cameras to further improve the performance in terms of number of social distancing breaches detected, accurately pursuing walking pedestrians etc. We highlight the performance benefits of our approach in different static and dynamic indoor scenarios. 
\end{abstract}


% \fi
\IEEEpeerreviewmaketitle

\ifCLASSOPTIONcompsoc
\IEEEraisesectionheading{\section{Introduction}\label{sec:introduction}}
\else
\section{Introduction}
\label{sec:introduction}
\fi

% Impacts of the Covid-19 pandemic 
\IEEEPARstart{T}{he} COVID-19 pandemic has caused significant disruption to daily life around the world. As of August 10, 2020, there have been $19.8$ million confirmed cases worldwide with more than $730$ thousand fatalities. Furthermore, this pandemic has caused significant economic and social impacts.

% Importance of social distancing
At the moment, one of the best ways to prevent contracting COVID-19 is to avoid being exposed to the coronavirus. Organizations such as the Centers for Disease Control and Prevention (CDC) have recommended many guidelines including maintaining social distancing, wearing masks or other facial coverings, and frequent hand washing to reduce the chances of contracting or spreading the virus. Broadly, social distancing refers to the measures taken to reduce the frequency of people coming into contact with others and to maintain at least 6 feet of distance between individuals who are not from the same household. Several groups have simulated the spread of the virus and shown that social distancing can significantly reduce the total number of infected cases~\cite{Mao2011}, \cite{pmid23763426}, \cite{pmid26847017}, \cite{pmid19104659}, \cite{pmid18401408}. 

\begin{figure}[t]
      \centering
      \includegraphics[width=\columnwidth,height=6cm]{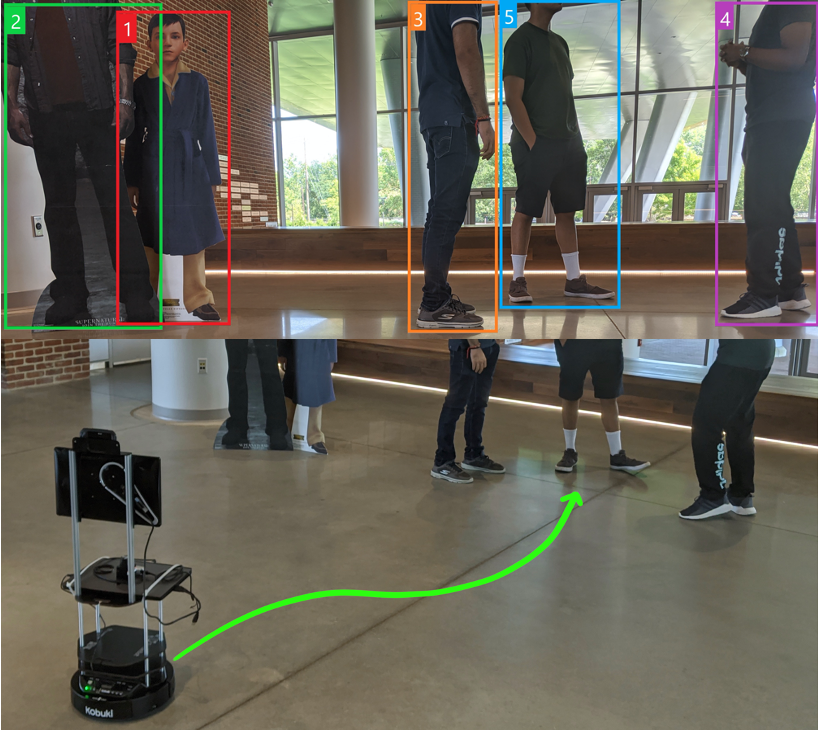}
      \caption {\small{Our robot detecting non-compliance to social distancing norms, classifying non-compliant pedestrians into groups and autonomously navigating to the group with currently the most people in it (a group with 3 people in this scenario). The robot encourages the non-compliant pedestrians to move apart and maintain at least 6 feet of social distance by displaying a message on the mounted screen. Our COVID-robot also captures thermal images of the scene and transmits them to appropriate security/healthcare personnel.}}
      \label{fig:cover-image}
      \vspace{-10pt}
\end{figure}

% Problems with maintaining/enforcing social distancing
A key issue is developing guidelines and methods to enforce these social distance constraints in public or private gatherings at indoor or outdoor locations. This gives rise to many challenges
including, framing reasonable rules that people can follow when they use public places such as supermarkets, pharmacies, railway and bus stations, spaces for recreation and essential work, and how people can be encouraged to follow the new rules. In addition, it is also crucial to detect when such rules are breached so that appropriate counter-measures can be employed. Detecting social distancing breaches could also help in contact tracing \cite{social-distance}. 

% Existing solutions for detecting crowding
Many technologies have been proposed for detecting excessive crowding or conducting contact tracing, and most of them use some form of communication.  Examples of this communication include WiFi, Bluetooth, tracking based on cellular connectivity, RFID, Ultra Wide Band (UWB) etc. Most of these technologies work well only in indoor scenes, though cellular have been used outdoors for tracking pedestrians. In addition, many of these technologies such as RFID, UWB, etc. require additional infrastructure or devices to track people indoors. In other cases, technologies such as WiFi and Bluetooth are useful in tracking only those people connected to the technologies using wearable devices or smartphones. This limits their usage for tracking crowds and social distancing norms in general environments or public places, and may hinder the use of any kind of counter-measures.

% Our solution
\noindent {\bf Main Results:} We present a vision-guided mobile robot (COVID-robot) to monitor scenarios with low or high-density crowds and prolonged contact between individuals. We use a state-of-the-art algorithm for autonomous collision-free navigation of the robot in arbitrary scenarios that uses a hybrid combination of a Deep Reinforcement Learning (DRL) method and traditional model-based method.  We use pedestrian detection and tracking algorithms to detect groups of people in the camera's Field Of View (FOV) that are closer than 6 feet from each other. Once social distance breaches are detected, the robot prioritizes groups based on their size, navigates to the largest group and encourages following of social distancing norms by displaying an alert message on a mounted screen. For mobile pedestrians who are non-compliant, the robot tracks and pursues them with warnings. 

Our COVID-robot uses inexpensive visual sensors such as an RGB-D camera and a 2-D lidar to navigate and to classify pedestrians that violate social distance constraints as \textit{non-compliant} pedestrians. In indoor scenarios, our COVID-robot uses the CCTV camera setup (if available) to further improve the detection accuracy and check a larger group of pedestrians for social distance constraint violations. We also use a thermal camera, mounted on the robot to wirelessly transmit thermal images. This could help detect persons who may have a high temperature without revealing their identities and protecting their private health information. 
 
 %identities and health data are kept private. Our system could ideally replace government/health personnel who enforce social distancing, measure temperatures in spaces such as airports, railway and bus stations, hospitals, grocery stores etc. This also protects such security and health personnel from over exposure to crowds and reduces the chances of them contracting the coronavirus.

% Main Contributions
% 1. Novel mobile robot system.
% 2. Integration with existing CCTV systems. Eliminates the need for extra infrastructure.
% 3. Novel, simple and fast (real-time) method to estimate distances between points on a "transformed" imaged based on Homography.
% 4. Novel grouping algorithm based on proximity between people and novel goal-selection based on a locked pedestrian.
% 5. An interface that communicates/encourages people to observe social distancing norms. 

\no Our main contributions in this work are: \\
\no \textbf{1.} A mobile robot system that detects breaches in social distancing norms, autonomously navigates towards groups of \textit{non-compliant} people, and encourages them to maintain at least 6 feet of distance. We demonstrate that our mobile robot monitoring system is effective in terms of detecting social distancing breaches in static indoor scenes and can enforce social distancing in all of the detected breaches. Furthermore, our method does not require the humans to wear any tracking or wearable devices.

\no \textbf{2.} We also integrate a CCTV setup in indoor scenes (if available) with the COVID-robot to further increase the area being monitored and improve the accuracy of tracking and pursuing dynamic non-compliant pedestrians. This hybrid combination of static mounted cameras and a mobile robot can further improve the number of breaches detected and enforcements by up to 100\%.

\no \textbf{3.} We present a novel real-time method to estimate distances between people in images captured using an RGB-D camera on the robot and CCTV camera using a homography transformation. The distance estimate has an average error of 0.3 feet in indoor environments.

\no \textbf{4.} We also present a novel algorithm for classifying non-compliant people into different groups and selecting a goal that makes the robot move to the vicinity of the largest group and enforce social distancing norms. 

\no \textbf{5.} We integrate a thermal camera with the robot and wirelessly transmit the thermal images to appropriate security/healthcare personnel. The robot does not record temperatures or perform any form of person recognition to protect people's privacy.    

% DO YOU MAKE SURE THAT THE ROBOT CAN NAVIGATE TO THE BEST LOCATION (e.g. IN FRONT OF PERSON) TO GET THE MOST ACCURATE TEMPERATURE READIND. WHAT ARE THE GUIDLINES AND HOW DO YOU ENSURE THE BEST POSITIONS This helps to detect individuals who might have a fever and later in contact tracing. 
% HOW HAVE YOU EVALUATED YOUR RESULTS; HOW MANY SCENARIOS, HOW DO YOU PERFORM EVALUTION

We have evaluated our method quantitatively in terms of accuracy of localizing a pedestrian, the number of social distancing breaches detected in static and mobile pedestrians, and our CCTV-robot hybrid system. We also measure the time duration for which the robot can track a dynamic pedestrian. Qualitatively, we highlight the trajectories of the robot pursuing dynamic pedestrians when using only its RGB-D sensor as compared to when both the CCTV and RGB-D cameras are used.

The rest of the paper is organized as follows. In Section 2, we present a brief review of related works on the importance of and emerging technologies for social distancing and robot navigation. In Section 3, we provide a background on robot navigation, collision avoidance, and pedestrian tracking methods used in our system.  We describe new algorithms used in our robot system related to grouping and goal-selection, CCTV setup, thermal camera integration, etc in Section 4. In Section 5, we evaluate our COVID-robot in different scenarios and demonstrate the effectiveness of our hybrid system (robot + CCTV) and compare it with cases where only the robot or a standard static CCTV camera system is used.

% \begin{figure}[t]
%       \centering
%       \includegraphics[width=\columnwidth,height=5.25cm]{Images/cat.jpg}
%       \caption {The hardware components of our system.}
%       \label{fig:turtlebot}
%       \vspace{-15pt}
% \end{figure}
\section{Related Works}
In this section, we review the relevant works that discuss the effectiveness of social distancing and the different technologies used to detect breaches of social distancing norms. We also give a brief overview of prior work on collision avoidance and pedestrian tracking. 

\subsection{Effectiveness of Social Distancing}
Works that have simulated the spread of a virus \cite{Mao2011}, \cite{pmid23763426}, \cite{pmid26847017}, \cite{pmid19104659}, \cite{pmid18401408} demonstrate different levels of effectiveness of different kinds of social distancing measures. Effectiveness of a social distancing measure is evaluated based on two factors: (1). the basic reproduction number $R_o$, and (2) the attack rate. $R_o$ is the average number of people to whom an infected person could spread the virus during the course of an outbreak. The attack rate is the ratio between the total number of infected cases over the entire course of the outbreak~\cite{social-distance}.

For instance, in \cite{Mao2011}, in a workplace setting, the attack rate can be reduced by up to $82\%$ if three consecutive days are removed from the workdays for $R_o = 1.4$ \cite{Mao2011}. Similarly, in an $R_o = 1.4$ setting, maintaining 6 feet or more between persons in the workplace could reduce the attack rate by up to $39.22\%$ \cite{pmid23763426} or reduce the rate by $11\%$ to $20\%$ depending on the frequency of contact with other employees \cite{pmid26847017}. Other works that have studied the effects of self-isolation \cite{pmid19104659}, \cite{pmid16642006}, show that it could reduce the peak attack rate by up to 89\% when $R_o < 1.9$. 

\subsection{Emerging Technologies for Social Distancing}
Recently, many techniques have been proposed to monitor whether
people are maintaining the 6-feet social distance. For instance, workers in Amazon warehouses are monitored for social distancing breaches using CCTV cameras \footnote{ Link:
\href{https://www.cnbc.com/2020/06/16/amazon-using-cameras-to-enforce-social-distancing-rules-at-warehouses.html}{Amazon CCTV}}. Other methods include using wearable alert devices \footnote{Link: \href{https://www.safespacer.net}{Wearable devices 1}, \href{https://spectrum.ieee.org/the-human-os/biomedical/devices/wearables-track-social-distancing-sick-employees-workplace}{2}, and \href{https://www.safeteams.co}{3}}. Such devices work using Bluetooth or UWB technologies. Companies such as Apple and Google are developing contact tracing applications that can alert users if they come in contact with a person who could be infected \footnote{\href{https://www.google.com/covid19/exposurenotifications/}{Google and Apple contact tracing}}.

A comprehensive survey of all the technologies that can be used to track people to detect if social distancing norms are followed properly is given in \cite{social-distance}. This includes a discussion of pros and cons of technologies such as WiFi, Zigbee, RFID, Cellular, Bluetooth, Computer Vision, AI, etc. However, almost all of these technologies require new static, indoor infrastructure such as WiFi routers, Bluetooth modules, central RFID hubs, etc. Technologies such as RFID and Zigbee  also require pedestrians to use wearable tags to localize them. 

Most of these technologies are also mostly limited to indoor scenes, with the exception of cellular-based tracking and do not help in \textit{reacting} to cases where people do not follow social distancing guidelines. In \cite{dog-robot}, a quadruped robot with multiple on-board cameras and a 3-D lidar is used to enforce social distancing in outdoor crowds using voice instructions. Our work is complimentary to these methods and also helps react to social distancing violations. Although we evaluate our system indoors, it was trivially be extended to outdoor scenes in the future.

\subsection{Collision-Free Navigation in Crowded Scenarios}
The problem of collision-free navigation has been extensively studied in robotics and related areas. Recently, some promising methods for navigation with noisy sensor data have been based on Deep Reinforcement Learning (DRL) methods~\cite{JHow1,JHow2}. These methods work well in the presence of sensor uncertainty and produce better empirical results when compared to traditional methods such as Velocity Obstacle-based methods \cite{RVO,ORCA}.  
These methods include training a decentralized collision avoidance policy by using only raw data from a 2-D lidar, the robot's odometry, and the relative goal location~\cite{JiaPan1}. The policy is extended by combining it with control strategies~\cite{JiaPan2}. Other works have developed learning-based policies that implicitly fuse data from multiple perception sensors to handle occluded spaces~\cite{crowdsteer} and to better handle the Freezing Robot Problem (FRP)~\cite{densecavoid}. Other hybrid learning and model-based methods include \cite{of-vo}, which predicts the pedestrian movement through optical flow estimation. \cite{frozone}  constructs a potential freezing zone that is used by the robot to prevent freezing and improve the pedestrian-friendliness of the robot's navigation. 
% Socially acceptable navigation \cite{socially-aware} and safety guarantees have been provided by identifying previously unseen scenarios and performing more cautious maneuvers \cite{JHow-uncertainty}, thus extending existing DRL-based methods. 
Our navigation approach is also based on DRL and can be combined with any of these methods.

\section{Background And Overview}
In this section, we provide a brief overview of the collision avoidance scheme used in our system, our pedestrian detection and tracking method, and our criteria for social distancing. 

\subsection{DRL-Based Collision Avoidance}
We use an end-to-end Deep Reinforcement Learning-based (DRL) policy \cite{JiaPan1} to generate collision-free velocities for the robot. We chose a DRL-based method because it performs well in the presence of sensor uncertainty, and have better empirical results than traditional collision avoidance methods. The collision avoidance policy is trained in a 2.5-D simulator with a reward function that (i) minimizes the robot's time to reach its goal, (ii) reduces oscillatory motions in the robot, (iii) heads towards the robot's goal, and most importantly, (iv) avoids collisions. At each time instance, the trained DRL policy $\pi_{\theta}$ takes 2-D lidar data observations ($\textbf{o}^t_{lidar}$), the relative goal location ($\textbf{o}^t_{goal}$), and the robot's current velocity ($\textbf{o}^t_{vel}$) as inputs to generate collision-free velocities $\textbf{v}^{DRL}$. Formally,

\begin{equation}
    \textbf{v}^{DRL} \sim \pi_{\theta}(\textbf{\textbf{v}}^t | \textbf{o}^t_{lidar}, \textbf{o}^t_{goal}, \textbf{o}^t_{vel}).
\end{equation}

This velocity is then post-processed using Frozone \cite{frozone} to eliminate velocities that lead to the Freezing Robot Problem (FRP).

\subsection{Frozone}
Frozone \cite{frozone} is a state-of-the-art collision avoidance method for navigation in moderate to dense crowds ($\ge 1 $ person$/m^2$) that uses an RGB-D camera to track and predict the future positions and orientations of pedestrians relative to the robot. Its primary focus is to simultaneously minimize the occurrence of FRP \cite{freezing1} and any obtrusion caused by the robot's navigation to nearby pedestrians. FRP is defined as any scenario where the robot's planner is unable to compute velocities that move the robot towards its goal.  When navigating among humans, the robot must ensure that it does not freeze, as it severely affects its navigation and causes inconvenience to the humans around it.

Frozone's two core ideas are as follows. The robot first classifies pedestrians into \textit{potentially freezing} (more probable of causing freezing) and \textit{non-freezing} pedestrians based on their walking speeds and directions by predicting their future positions over a time horizon. The robot then constructs and avoids a spatial region called the \textit{Potential Freezing Zone} (PFZ). The PFZ corresponds to the set of locations where the robot has the maximum probability of freezing and being obstructive to the pedestrians around it. Formally, the PFZ is constructed as follows:

\begin{equation}
    PFZ = Convex Hull (\hat{\textbf{p}}^{ped}_i), \quad i \in {1,2,...,K.},
\end{equation}

\no where $\hat{\textbf{p}}^{ped}_i$ is the predicted future position of the $i^{th}$ pedestrian, and K is the total number of potentially freezing pedestrians. $\hat{\textbf{p}}^{ped}_i$ is calculated as, $\hat{\textbf{p}}^{ped}_i = \textbf{p}^{ped}_i + \textbf{v}^{ped}_i\Delta t, where \quad i \in {1,2,...,K}$.

The symbols $\textbf{p}^{ped}_i$ and $\textbf{v}^{ped}_i$ denote the $i^{th}$ pedestrian's current position and velocity vectors relative to the robot, and $\Delta t$ is the time horizon over which prediction is done. If the distance between the robot and the closest potentially freezing pedestrian is less than a threshold distance, the robot deviates its current velocity direction (computed by the DRL method) away from the PFZ. 

% Can add background for deviation angle computation if necessary.

\subsection{Pedestrian Detection and Tracking} \label{ped-detect}
A lot of work has been done on object detection and tracking in recent years, especially on methods based on deep learning. For detecting and tracking pedestrians, we use the work done in \cite{ped-detect-track} based on Yolov3 \cite{YOLOv3}, a scheme that achieves a good balance between speed and tracking accuracy. The input to the tracking scheme is an RGB image and the output is a set of bounding box coordinates for all the pedestrians detected in the image. 

The bounding boxes are denoted as $\mc{B}= \{ \bb{B}_{k} \ | \ \bb{B} = [\textrm{top left}, m_{\bb{B}}, n_{\bb{B}}],  \in \mc{H} \}$, where $\mc{H}$ is the set of all pedestrian detections, $\textrm{top left}, m_{\bb{B}},$ and $n_{\bb{B}}$ denote the top left corner coordinates, width, and height of the $k^{th}$ bounding box $\bb{B}_k$, respectively. Apart from these values, Yolov3 also outputs a unique ID for every person in the RGB image, which remains constant as long as the person remains in the camera's FOV. Since Yolov3 requires RGB images, the images from both the RGB-D and the CCTV cameras can be used for detecting pedestrians.

% This section needs more attention. 
% Add figure to show scenarios 

\begin{figure}[t]
      \centering
      \includegraphics[width=\columnwidth,height=4.5cm]{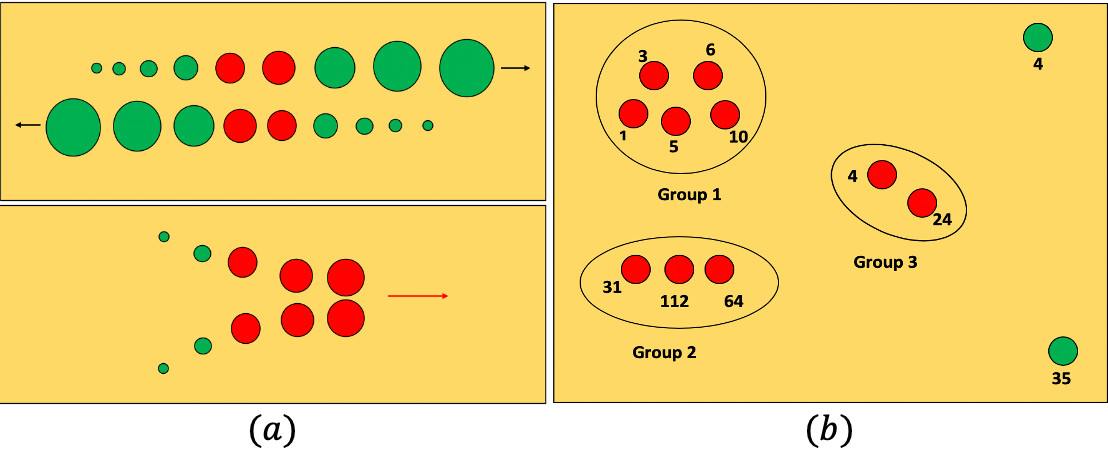}
      \caption {\small{\textbf{a.} The criteria used to detect whether two pedestrians violate the social distance constraint.  This figure shows two pedestrians represented as circles in two different scenarios. The increasing size of the circles denotes the passage of time. The green circles represent time instances where the pedestrians maintained $> 6$ feet distance, and the red circles represent instances where they were closer than 6 feet. \textbf{Top:} Two pedestrians passing each other.  This scenario is not reported as a breach since the duration of the breach is short. \textbf{Bottom:} Two pedestrians meeting and walking together. This scenario is reported as a breach of social distancing norms. \textbf{b.} A top-down view of how non-compliant pedestrians (denoted as red circles) are classified into groups. The numbers beside the circles represent the IDs of the pedestrians outputted by Yolov3. The compliant pedestrians (green circles) are not classified into groups as the robot does not have to encourage them to maintain the appropriate social distance. In the scenario shown, the robot would first attend to Group 1.}}
      \label{fig:breach-criteria}
      \vspace{-15pt}
\end{figure}

\subsection{Criteria for Social Distancing Breach} \label{breach-criteria}
% Breaching social distancing norms could mean several things such as not obeying stay-at-home orders or attending a populous event or in general increasing the frequency of coming into contact with others. 
We mainly focus on detecting scenarios where individuals do not maintain a distance of at least 6 feet from others \textit{for a given period of time} (we choose a 5-second threshold). We choose to detect this scenario because it is a fundamental social distancing norm during all stages of a pandemic, even as people begin to use public spaces and restrictions are lifted. 

% Challenges in detecting this scenario
An important challenge in detecting when individuals are not maintaining appropriate distances amongst themselves is avoiding false negatives. For example, two or more people passing each other should not be considered a breach, even if the distance between them was less than 6 feet for a few moments (see Figure \ref{fig:breach-criteria}a). Another challenge is detecting pedestrians and estimating the distances between them in the presence of occlusions. This can be addressed in indoor scenarios by using available static mounted CCTV cameras. 

% THERMAL SENSORS: HOW DO THEY WORK? HOW CLOSE SHOULD THEY BE PLACED IN FRONT OF THE HUMANS AND WHAT ORIENTATIONS> HOW DOES THE ACCURACY VARIES

\section{Our Method}
\begin{figure}[t]
      \centering
      \includegraphics[width=\columnwidth,height=5.25cm]{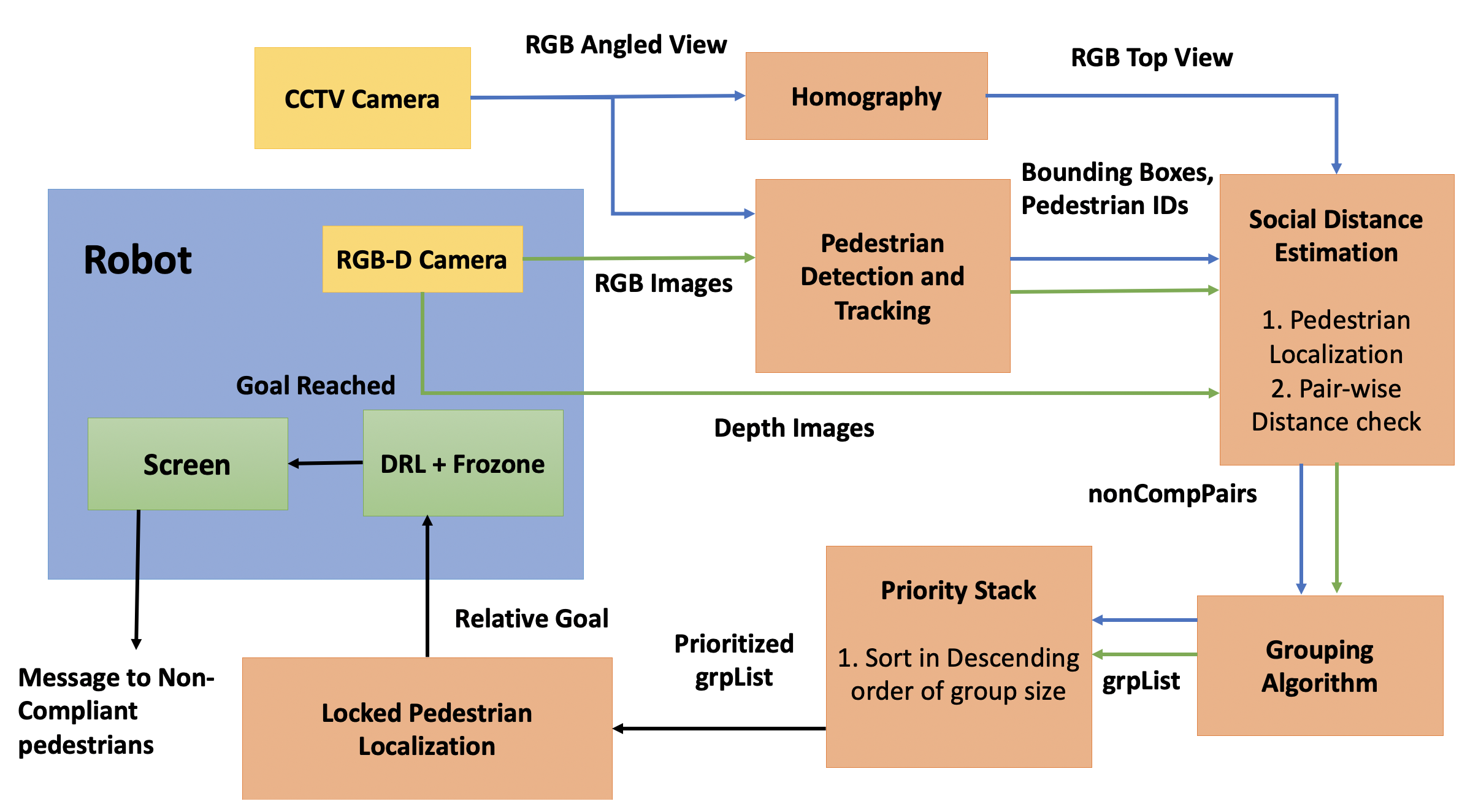}
      \caption {\small{Overall architecture of COVID-Robot and social distance monitoring: Our method's main components are (i) Pedestrian tracking and localization, (ii) Pairwise distance estimation between pedestrians, (iii) Classifying pedestrians into groups, (iv) Selecting a locked pedestrian in the largest group, (v) Using a hybrid collision avoidance method to navigate towards the locked pedestrian, and (vi) Display an alert message to the non-compliant pedestrians encouraging them to move apart. }}
      \label{fig:system-arch}
    %   \vspace{-15pt}
\end{figure}

In this section, we first discuss how our method effectively detects a breach in social distancing norms. We refer to people who violate social distancing norms as \textit{non-compliant} pedestrians. We then describe how we classify non-compliant pedestrians into groups and compute the goal for the robot's navigation based on the size of each group. Our overall system architecture is shown in figure \ref{fig:system-arch}.

% SHOW SOME 2D SCENARIOS WHERE PEDESTRIANS ARE SHOWN AS DOTS, AND HOW YOU CHEKC FOR SOCIAL DISTANCE BETWEEN THEM. 
\subsection{Breach Detection} \label{breach-detection}
As mentioned in Section \ref{breach-criteria}, if certain individuals do not maintain a distance of at least 6 feet from each other, the system must report a breach. The robot's on-board RGB-D camera and the CCTV camera setup (whenever available) continuously monitor the states of individuals within their sensing range. At any instant, breaches could be detected by the robot's RGB-D camera and/or the CCTV camera.   

\subsubsection{Social Distance Estimation Using RGB-D Camera} \label{ped-localization-rgbd}
\begin{figure}[t]
      \centering
      \includegraphics[width = \columnwidth, height = 1.8in]{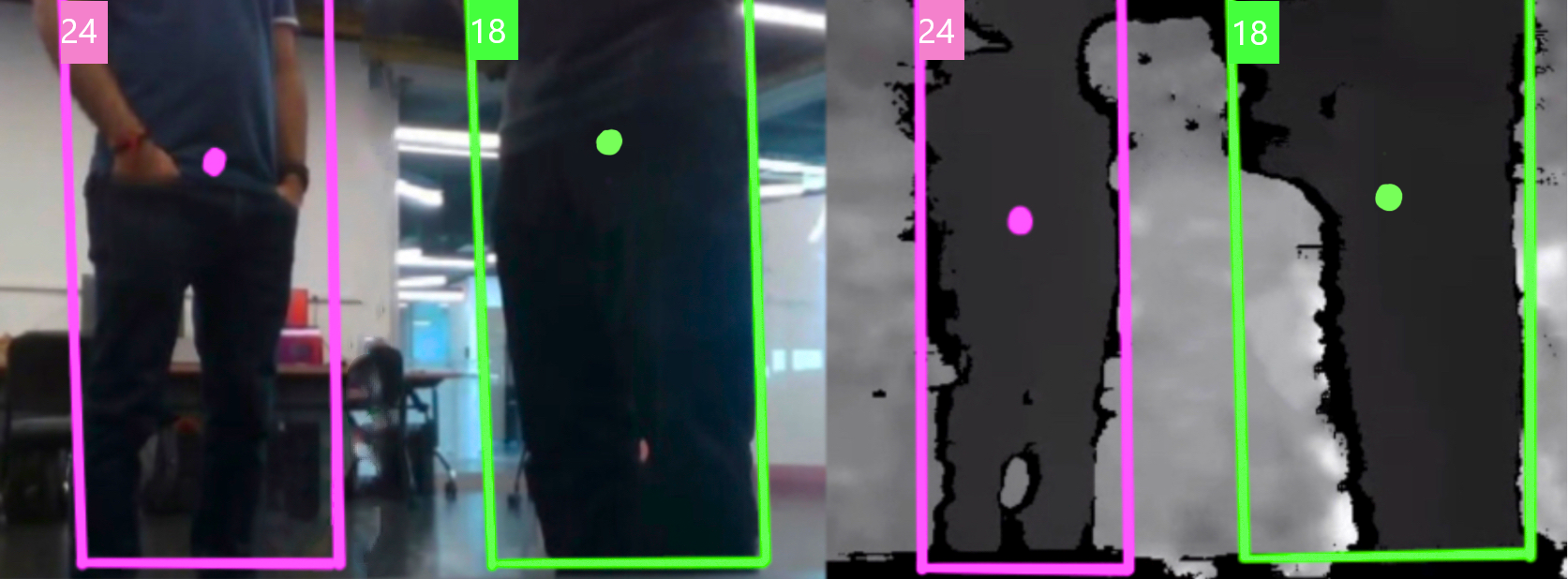}
      \caption {\small{\textbf{Left:} Two pedestrians detected in the RGB image of the robot's RGB-D camera with the bounding box centroids marked in pink and green. \textbf{Right:} The same bounding boxes superimposed over the depth image from the RGB-D camera. The pedestrians are localized and the distance between them is estimated by the method detailed in Section \ref{ped-localization-rgbd}. 
    %   \textbf{b.} Two pedestrians detected in the original angled view from the CCTV camera with the point corresponding to their feet marked in green. The homography rectangle is marked in red and its corner points are numbered. \textbf{c.} The top view image of the homography rectangle obtained after applying the homography transformation. A coordinate frame (shown in green) is fixed with corner point 1 of the homography rectangle as its origin.
      }}
      \label{fig:social-dist-estimation}
    %   \vspace{-15pt}
\end{figure}

We first describe how we localize a person detected in the RGB image (Section \ref{ped-detect}) with respect to the robot by using its corresponding depth image from the RGB-D camera. The depth and RGB images from the RGB-D camera have the same widths and heights and are aligned by default to be looking at the same subjects (see figure \ref{fig:social-dist-estimation}). We denote the depth image at any time instant \textit{t} as $I^t$, and the value contained in a pixel at coordinates $(i, j)$  is the proximity of an object at that part of the image. 

Formally, $I^t = \{C \in \mathbb{R}^{h \times w} : f < C_{ij} < R,$ and $1 \le i \le w \qquad \text{and} \qquad 1 \le j \le h.$  Here, f is an offset distance from the RGB-D camera from where depth can be accurately measured, and R is the maximum range in which depth can be measured. Symbols \textit{w}, \textit{h}, \textit{i}, and \textit{j} represent the image's width, height, and the indices along the width and height, respectively. Using this data, we localize a detected pedestrian \textit{P} as follows.  

First, the detection bounding boxes from the RGB image are superimposed over the depth image. Next, the minimum 10\% of the pixel values inside the bounding box $\bb{B}_P$ are averaged to obtain the mean distance ($d_{avg}$) of pedestrian \textit{P} from the camera. Denoting the centroid of the bounding box $\bb{B}_P$ as $[x^{\bb{B}_P}_{cen}, y^{\bb{B}_P}_{cen}]$, the angular displacement $\psi_P$ of the pedestrian relative to the robot can be computed as:
\begin{equation}
    \psi_P = \left(\frac{\frac{w}{2} - x^{\bb{B}_P}_{cen}}{w}\right) * FOV_{cam},
    \label{eqn:psi-angle-calculation}
\end{equation}
\no where $FOV_{cam}$ is the field of view angle of the camera. This calculates the angle in a coordinate system attached to the robot such that its X-axis is along the robot's forward direction and Y-axis is towards the robot's left. $\psi_P$ can range between $[-\frac{FOV_{cam}}{2}, \frac{FOV_{cam}}{2}]$. The pedestrian's position with respect to the robot is then calculated as $[p^{P}_x, p^{P}_y]$ = $d_{avg}$ * [$\cos{\psi_P}, \sin{\psi_P}$].

To estimate the distances between a pair of pedestrians, say $P_a$ and $P_b$, we use the Euclidean distance function given by,

\begin{equation}
    dist(P_a, P_b) = \sqrt{(p^{P_a}_x - p^{P_b}_x) + (p^{P_a}_y - p^{P_b}_y).}
    \label{dist-measure}
\end{equation}

\no If $dist(P_a, P_b) < 6$ feet for a period of time T (we choose 5 seconds), then the robot reports a breach for that pair of individuals. This process is repeated in a pairwise manner for all the detected individuals or pedestrian, and a list of pairs of non-compliant pedestrian IDs is obtained from the sensor data. 

% \begin{figure}[t]
%       \centering
%       \includegraphics[width=\columnwidth,height=5.25cm]{Images/cat.jpg}
%       \caption {Figures showing distance estimated between individuals in RGB-D and CCTV scenario (show homography).}
%       \label{fig:social-dist-estimation}
%     %   \vspace{-15pt}
% \end{figure}

\subsubsection{Social Distance Estimation Using a CCTV Camera} \label{ped-localization-cctv}
While the robot's RGB-D camera has the advantage of being mobile and being able to detect breaches anywhere, it is limited by a small FOV and sensing range. If a breach of social distancing occurs outside this sensing range, it will not be reported. To mitigate this limitation, we utilize an existing CCTV camera setup in indoor settings to widen the scope for detecting breaches. Pedestrian detection and tracking are done as described in Section \ref{ped-detect}. We estimate distances between individuals as follows.

\begin{figure*}[t]
      \centering
      \includegraphics[width=\textwidth,height=5.25cm]{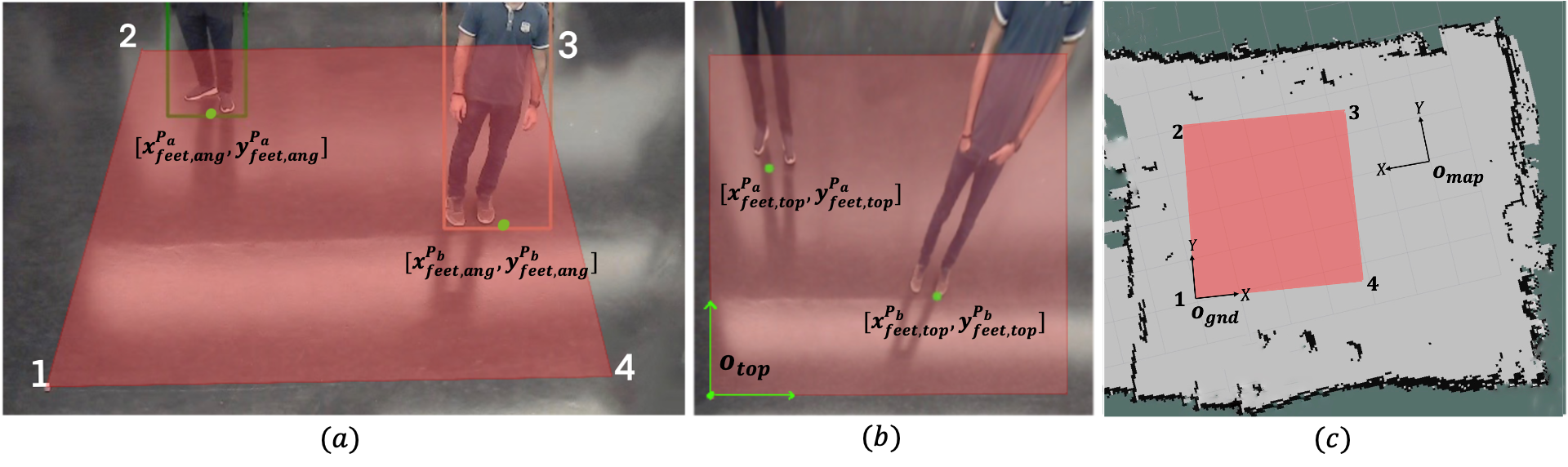}
      \caption {\textbf{a.} The angled view of the homography rectangle marked in red and corners numbered from the CCTV camera. The green dots mark the points corresponding to a person's feet in this view. \textbf{b.} The top view of the homography rectangle after transformation and the origin of the top view coordinate system is marked as $o
      _{top}$. The coordinates of the feet points are also transformed using the homography matrix.  \textbf{c.} A map of the robot's environment with free space denoted in gray and obstacles denoted in black with a coordinate frame at origin $o_{map}$. The homography rectangle is marked in red and the ground plane coordinate system is shown with the origin $o_{gnd}$. }
      \label{fig:3-transformations}
    %   \vspace{-15pt}
\end{figure*}

\textbf{Homography:} All CCTV cameras are mounted such that they provide an \textit{angled} view of the ground plane. However, to accurately calculate the distance between any two pedestrians on the ground, a top view of the ground plane is preferable. To obtain the top view, we transform the CCTV camera's angled view of the ground plane by applying a homography transformation to four points on the ground plane in the angled view. The four points are selected such that they form the corners of the maximum area of a rectangle that can fit within the FOV of the CCTV camera (see Figure \ref{fig:3-transformations}a and b). Let us call this rectangle the homography rectangle. The four points are transformed as,

% The four points look like the corner points of trapezoid in the CCTV perspective.
% Can we add the OpenCV link as a reference?

\begin{equation}
    \begin{bmatrix}
    x_{corn, top} \\
    y_{corn, top} \\
    1
    \end{bmatrix}
    = M * \begin{bmatrix}
    x_{corn, ang} \\
    y_{corn, ang} \\
    1
    \end{bmatrix}
    \label{homography}
\end{equation}

\no where $x_{corn, ang}$ and $y_{corn, ang}$ denote the pixel coordinates of one of the four points in the angled CCTV view image. $x_{corn, top}$ and $y_{corn, top}$ denote the same point after being transformed to the top view, and M is the scaled homography matrix. The homography matrix is computed using standard OpenCV functions.

\textbf{Distance Estimation Between Pedestrians:} After obtaining the homography matrix, we localize each detected pedestrian within the homography rectangle as follows. We first obtain a point corresponding to the feet of a pedestrian P ([$x^P_{feet, ang}, y^P_{feet, ang}$]) by averaging the coordinates of the bottom left and the bottom right corners of the bounding box of the pedestrian (see Figure \ref{fig:3-transformations}a) in the angled CCTV view. This point is then transformed to the top view using Equation \ref{homography} as $[x^P_{feet, top}, y^P_{feet, top}]^T = M*[x^P_{feet, ang}, y^P_{feet, ang}]^T$ (see Figure \ref{fig:3-transformations}b). 

The distance between any two pedestrians $P_a$ and $P_b$ is first calculated by using Equation \ref{dist-measure} with the coordinates [$x^{P_a}_{feet, top}, y^{P_a}_{feet, top}$] and [$x^{P_b}_{feet, top}, y^{P_b}_{feet, top}$]. This distance is then scaled by an appropriate factor S to obtain the real-world distance between the pedestrians. The scaling factor is found by measuring the number of pixels in the image that constitute 1 meter in the real-world.

If the real-world distance between a pair of pedestrians is less than 6 feet for a period of time T, a breach is reported for that pair. A list of all the pairs of non-compliant pedestrian IDs is then obtained.

\subsection{Enforcing Social Distancing}
Once a breach is detected through the robot's RGB-D camera and/or the CCTV camera, the robot must navigate towards the location of the breach and encourages the non-compliant pedestrians to move away from each other through an alert message. If the non-compliant pedestrians are walking, the robot pursues them until they observe social distancing. Prior to this, the robot must compute the location of the breach relative to itself. We detail this process in the following sections. 

\subsubsection{Classifying People into Groups}

% \begin{figure}[t]
%       \centering
%       \includegraphics[width=\columnwidth,height=5.25cm]{Images/Grouping.png}
%       \caption {\small{A top-down view of how non-compliant pedestrians (denoted as red circles) are classified into groups. The numbers beside the circles represent the IDs of the pedestrians outputted by Yolov3. The compliant pedestrians (green circles) are not classified into groups as the robot does not have to encourage them to maintain the appropriate social distance. In the scenario shown, the robot would first attend to Group 1.}}
%       \label{fig:group-classification}
%     %   \vspace{-15pt}
% \end{figure}

In social scenarios, people naturally tend to walk or stand in groups. We define a group as a set of people who are closer than 6 feet from each other (see Figure \ref{fig:breach-criteria}b). Therefore, if the robot attends to a group, it can convey the alert message to observe social distancing to all the individuals in that group. In addition, when there are multiple groups of people breaching the social distancing norms, the robot can prioritize attending to each group based on the number of people in it. We classify non-compliant people into groups based on Algorithm \ref{algo:group-classification}. 

\begin{algorithm}[h]
\DontPrintSemicolon
  
  \KwInput{A list nonCompPairs of length $S_{input}$}
  \KwOutput{A list grpList}
%   \KwData{Testing set $x$}
%   $\sum_{i=1}^{\infty} := 0$ \tcp*{this is a comment}
%   \tcc{Now this is an if...else conditional loop}
  nonCompPairs $\gets $ List of pedestrian ID pairs breaching social distancing \;
  grpList $\gets$ nonCompPairs[0] \;    
  \For{i \text{from} 1 to $S_{input}$}    
        { 
        	counter $\gets 0$ \;
        	
        	\For{j \text{from} 0 to len(grpList)} 
                { \tcp*{len() returns the length of the list}
                intersection $\gets$ grpList[j] $\cap$ nonCompPairs[i] \;
                \If{intersection $\ne \emptyset$} 
                { grpList[j] $\gets$ grpList[j] $\cup$ grpList[i] \;
                }
                \Else
                { counter $\gets$ counter + 1 \;
                }
                }
            \If{counter $==$ len(grpList)} 
            { grpList.append(nonCompPairs[i])
            }
        }
%   \KwRet grpList        

\caption{Group Classification Algorithm.}
\label{algo:group-classification}
\end{algorithm}

In Algorithm \ref{algo:group-classification}, nonCompPairs is a list that contains the IDs of all the pairs of non-compliant pedestrians obtained in Section \ref{breach-detection}. grpList is a list of groups where each group contains the IDs of people who have been assigned to it. For example, if nonCompPairs containing pedestrian IDs 1 to 5 looks like [(1, 2), (1,3), (2,3), (4,5)], then grpList would contain two groups and look like [\{1, 2, 3\}, \{4, 5\}]. Once the number of groups and the number of people in each group are known, the robot \textit{locks} a pedestrian in the group with the most people and navigates towards him/her.

\subsubsection{Locked Pedestrian}
Consider a dynamic group of non-compliant pedestrians. The robot's RGB-D camera or the CCTV camera must be able to track at least one member of that group to efficiently guide the robot towards that group. Our method chooses a person who has the least probability of moving out of the FOV of either the robot's RGB-D camera or the CCTV camera (depending on which camera detected the group), and \textit{locks} on to him/her. This person is called the \textit{locked pedestrian}. The identity of the locked pedestrian is updated as people's positions change. To find the locked pedestrian, we consider the centroid of the bounding box of each person in the largest group. The person whose centroid has the least lateral distance from the center of the image is chosen as the locked pedestrian. That is, the condition for locking a pedestrian is,

\begin{equation}
    x^{P_{lp}}_{cent} - \frac{w}{2} = \min_{k \in \mathcal{I}_{\mathcal{G}}} x^{P_k}_{cent} - \frac{w}{2}, 
\end{equation}
where $\mathcal{I}_{\mathcal{G}}$ is the set of IDs for the detected pedestrians in the current largest group and $P_{lp}$ denotes the locked pedestrian.

\subsubsection{Computing Goal Position Using an RGB-D Camera}
Once a pedestrian is locked, the robot localizes him/her relative to itself using the $d_{avg}$ and equation \ref{eqn:psi-angle-calculation} in Section \ref{ped-localization-rgbd}). That is,

\begin{equation}
    o^t_{goal} = d^{lp}_{avg} * [\cos{\psi_{P_{lp}}} \sin{\psi_{P_{lp}}}]^T,
\end{equation}

Where, $o^t_{goal}$ is the location of the goal relative to the robot, $d^{lp}_{avg}$ is the average distance and $\psi_{P_{lp}}$ is the angular displacement of the locked pedestrian from the robot respectively. The DRL method and Frozone use $o^t_{goal}$ to navigate the robot towards the locked pedestrian in a pedestrian friendly way without freezing.   

\subsubsection{Computing Goal Position Using a CCTV Camera}
If the CCTV camera detects a breach and a locked pedestrian, the goal computation for the robot requires homogeneous transformations between three coordinate frames: 1. the top-view image obtained after homography, 2. the ground plane, and 3. a map of the environment with which the robot is localized. These three coordinates with origins $o_{top}$, $o_{gnd}$ and $o_{map}$ respectively are shown in figure \ref{fig:3-transformations}b and c. First, the locked pedestrian's location in the ground plane coordinate frame is obtained. This is done as follows. 

Let us consider the point in the top-view image corresponding to the feet of the locked pedestrian $[x^{P_{lp}}_{feet, top}, y^{P_{lp}}_{feet, top}]$ and corner point 1 (see figure \ref{fig:3-transformations}) for the homography rectangle $[x_{corn, top}, y_{corn, top}] = o_{top}$. The angle between the two points in the image is calculated as, 

\begin{equation}
    \theta_{lp-corn, top} = \taninv(\frac{(y_{corn, top} - y^{P_{lp}}_{feet, top})}{(x_{corn, top} - x^{P_{lp}}_{feet, top})}). \\
    \label{angle-measure}
\end{equation}

We consider corner point 1 of the homography rectangle in the real world to be the origin of the coordinate system fixed to the ground plane ($o_{gnd}$) with its X and Y axes aligned with the X and Y axes of the top view image (see Figures \ref{fig:3-transformations}a and c. Therefore, the angle $\theta_{lp-corn, top}$ also corresponds to the angle between the two points on the ground plane $\theta_{lp-corn, gnd}$. The Euclidean distance ($r_{lp-corn, top}$) between the points on the top view image is calculated using Equation \ref{dist-measure}. The real world distance between the points, denoted as $r_{lp-corn, gnd}$ is then obtained by scaling $r_{lp-corn, top}$ using the factor S. 

% \begin{figure*}[t]
%       \centering
%       \includegraphics[width=\textwidth,height=5.25cm]{Images/Coordinate_Frames2.png}
%       \caption {A map of the robot's environment. The homography rectangle is marked in red and the ground plane coordinate system is shown with origin $o_{gnd}$. The map coordinate frame is shown with origin $o_{map}$.}
%       \label{fig:3-transformations}
%     %   \vspace{-15pt}
% \end{figure*}

The location of the locked pedestrian in the ground coordinate frame is calculated as $[x^{P_{lp}}_{feet, gnd} y^{P_{lp}}_{feet, gnd}]^T = o_{gnd}^T + r_{lp-corn, gnd}*[\cos\theta_{lp-corn, gnd} \,\,\, \sin\theta_{lp-corn, gnd}]^T$. 
 
The locked pedestrian's location is then converted to the map coordinate frame using a homogeneous transformation matrix as, $[x^{P_{lp}}_{feet, map}  y^{P_{lp}}_{feet, map}]^T = H^{map}_{gnd} * [x^{P_{lp}}_{feet, gnd}  y^{P_{lp}}_{feet, gnd}]^T$. Now, since the robot and the locked pedestrian are localized with respect to the map coordinate frame, the relative goal location for the robot can be obtained as follows,

\begin{equation}
    o^t_{goal} = [x^{P_{lp}}_{feet, map}  y^{P_{lp}}_{feet, map}]^T - [x_{robot, map} y_{robot, map}]^T.
\end{equation}

Where $x_{robot, map}$ and $y_{robot_map}$ are the X and Y coordinates of the robot in the map coordinate frame. Using this $o^t_{goal}$, the trained DRL policy computes the collision-free velocity towards the locked pedestrian.

\subsubsection{Multiple Groups and Lawnmower Inspection}
So far, we have discussed how a breach of social distancing norms can be detected using either the robot's RGB-D camera or an existing, independent CCTV camera setup. In the case where both cameras detect several groups of non-compliant pedestrians, the robot attends to the group with the most number of individuals. The robot attends to a group until everyone in the group observes the appropriate distancing measures. Once the robot is done attending to a group, the next largest group is selected and attended to. If the same group is detected in both cameras, the goal data computed using the CCTV camera will be used to guide the robot.

% WHY IS THE ISSUE OF LAWNMOWER INSPECTION IMPORTANT?
To improve the effectiveness of the integrated robot and CCTV system in detecting new non-compliant groups of pedestrians, the robot inspects the blind spots of the CCTV camera continuously by following the well-known lawnmower strategy. This expands the total area that the system is monitoring at any time instant. In addition, the lawnmower strategy guarantees that 100\% of an environment can be covered by navigating to a few fixed waypoints, although it does not guarantee an increase in the number of breaches detected.  
\subsection{Alerting and Encouraging Pedestrians}
Once the robot reaches the vicinity of the locked pedestrian, the robot first displays the reason why they were approached on its mounted screen; the estimated distance between the people in the group. The robot then displays a message encouraging the people to stay apart from each other. 

While this is a simplistic approach, this setup can be easily improved with a number of extensions in the future. For instance, the robot can also \textit{talk} to the people in a group by either playing a recorded message, or a message from the security authorities. It can also be extended to include virtual AI applications that can assist people by understanding the context of the scenario.

\subsection{Thermal Camera}
\begin{figure}[t]
      \centering
      \includegraphics[width=\columnwidth,height=5.50cm]{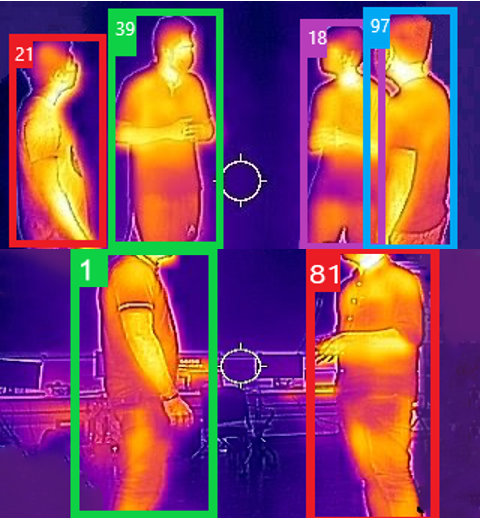}
      \caption {\small{Thermal images generated by the thermal camera that is wirelessly transmitted to appropriate security/healthcare personnel. The temperature signatures of the people irrespective of their orientations. We intentionally have a human in the loop to monitor people's temperature signatures, and we do not perform any form of facial recognition on people to protect their privacy. Pedestrians are detected on the thermal image to aid the personnel responsible for monitoring the area.}}
      \label{fig:thermal-cam}
    %   \vspace{-15pt}
\end{figure}

As mentioned previously, the robot is also equipped with a thermal camera that generates images based on the differences in temperatures of different regions that it observes (see Figure \ref{fig:thermal-cam}). Our pedestrian detection scheme detects and tracks people on these images and the results are then sent to appropriate security or healthcare personnel who detects if an individual's temperature signature is higher than normal. Measures can then be initiated to trace the person for future contact tracing. We intentionally choose to have a human in the loop instead of performing any form of facial recognition to protect people's privacy. 

Such a system would be useful in places where people's temperatures are already measured by security/healthcare personnel such as airports, hospitals etc. Monitoring people's temperatures remotely reduces exposure for security/healthcare personnel, thus reducing their risk of contracting the coronavirus.   

% WOULD YOU ADD A SUBSECTION ON HOW DO YOU PROVIDE WARNING TO PEDESTRIANS THAT VIOLATE? SOME INITIAL IDEAS AND MORE WORK LATER>

\section{Results and Evaluations}
In this section, we elaborate on how our system was implemented on a robot, explain the metrics we use to evaluate our system and analyze the effectiveness and the limitations of our method.

\subsection{Implementation}
% Should we mention research restrictions during Covid? 
We implement our method on a Turtlebot 2 robot customized with additional aluminium rods to attach a 15-inch screen to display messages to the non-compliant pedestrians. We specifically chose the Turtlebot 2 due to its ease of customization and its light-weight and tall structure. The pedestrian detection and tracking algorithm is executed on a laptop with an Intel i9 8th generation CPU and an Nvidia RTX2080 GPU mounted on the robot. We use an Intel Realsense (with $70^o$ FOV) RGB-D camera to sense pedestrians and a Hokuyo 2-D lidar ($240^o FOV$) to sense other environmental obstacles. 

hTo emulate a CCTV camera setup, we used a simple RGB webcam with a 1080p resolution mounted at an elevation. To process the images from the CCTV camera, we use a laptop with an Intel i7 7th generation CPU and an Nvidia GTX1060 GPU. We use a FLIR C3 thermal camera to generate the temperature signatures of the robot's surroundings. The ROS package for adaptive Monte-Carlo localization is used for locating the robot relative to the map coordinate frame.

% Add about thermal camera specs, packages used for localization

\subsection{Metrics}
We use the following metrics to evaluate our method. 

\begin{itemize}

    \item \textbf{Accuracy of pedestrian localization:} We compare the ground truth location of a pedestrian with the location estimated using our method as detailed in Sections \ref{ped-localization-rgbd} and \ref{ped-localization-cctv}. Higher localization accuracy translates to more accurate distance estimation and goal selection for the robot's navigation. 

    \item \textbf{Number of breaches detected:} This is the total number of locations in an environment at which a social distancing breach can be detected, given a total number of locations uniformly sampled from the environment. We measure this metric both in the presence and absence of occlusions in the environment. This metric provides a sense of the area in the environment that can be monitored by our system at any time instant. Higher values are better.
    
    \item \textbf{Number of enforcements:} The number of times the robot attended to a breach once it was detected. We again measure this in the presence and absence of occlusions in the environment. Ideally should be equal to the number of breaches detected.  
    
    \item \textbf{Tracking Duration for a mobile pedestrian:} We measure the time for which the robot is able to track a walking pedestrian. Since the robot's RGB-D camera has a limited FOV, the robot must rotate itself to track a pedestrian for a longer time. This metric is a measure of the robot's effectiveness in pursuing a mobile locked pedestrian (in a group of people who are walking together). 
    
\end{itemize}

\subsection{Experiments and Analysis}
\subsubsection{Accuracy of Pedestrian Localization} 
\begin{figure*}[t]

      \centering
      \includegraphics[width=\textwidth,height=5.5cm]{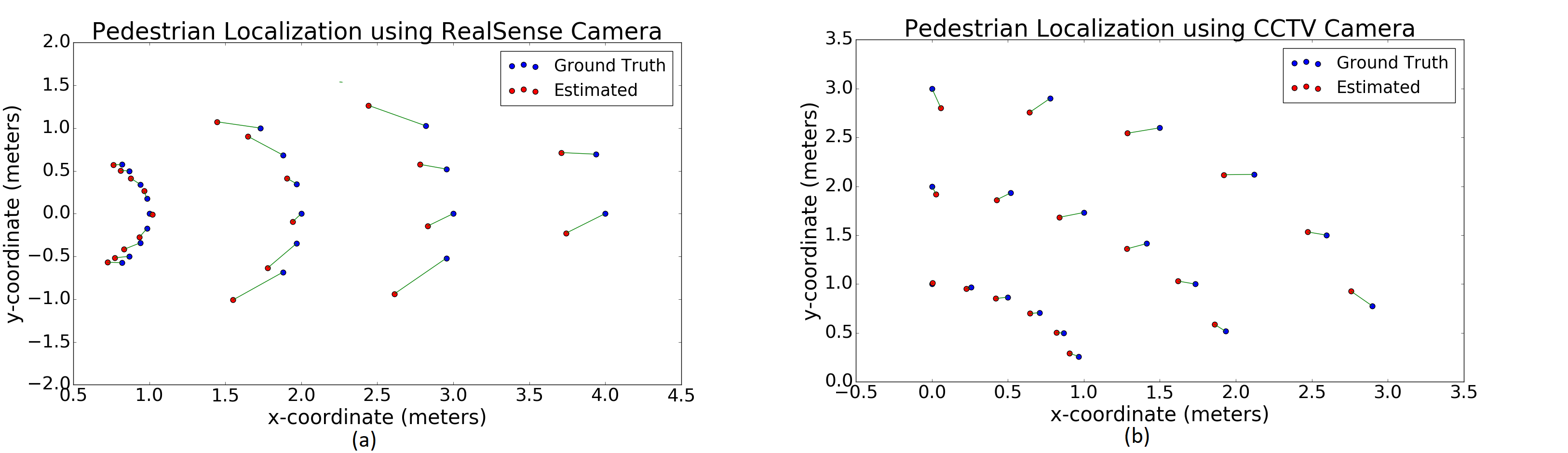}
      \caption {\small{Plots of ground truth (blue dots) versus pedestrian localization (red dots) when using the robot's Realsense camera and the static CCTV camera with more FOV. \textbf{a.} The estimates from the Realsense camera tend to have slightly higher errors because we localize pedestrians using averaged proximity values within their detection bounding boxes, which is affected by the size of the bounding boxes. \textbf{b.} Localization using the data from the CCTV camera is more accurate as it tracks a person's feet. This method is not affected by a person's orientation. We observe that in both cases, the localization errors are within the acceptable range of 0.3 meters. }}
      \label{fig:localization-accuracy}
    %   \vspace{-15pt}
\end{figure*}

We perform two sets of comparisons of the ground truth locations versus the estimated pedestrian location using 1. the robot's RGB-D camera, and 2. the CCTV setup. The plots are shown in Figure \ref{fig:localization-accuracy}, with the ground truth locations plotted as green circles and the estimated locations plotted as blue circles. The plot in Figure \ref{fig:localization-accuracy}a shows the pedestrian being localized with respect to a coordinate axis fixed to the robot, with its positive X-axis pointing in the robot's forward direction and the positive Y-axis pointing towards the robot's left. Figure \ref{fig:localization-accuracy}b shows a pedestrian being localized in the ground coordinate frame.

% RGB-D camera
We observe in Figure \ref{fig:localization-accuracy}a that when a pedestrian is closer to the robot and closer to the X-axis of the robot, the localization estimates closely match the ground truth. If a pedestrian is farther away from the robot or near the exterior limits of the RGB-D camera's FOV, the errors between the estimates and the ground truth values increase. This is mainly because the robot localizes a pedestrian based on the centroid of the bounding box of the pedestrian, which is located on the person's torso, whereas the ground truth is measured as a point on the ground. 

In addition, the orientation of the pedestrian relative to the RGB-D camera also affects the centroid of the bounding box and the localization estimate. However, since the maximum error between the ground truth and estimated values is within 0.3 meters, its effect on the social distance calculation and goal selection for the robot is within an acceptable limit. The accuracy can also be improved with higher FOV depth cameras in the future. 

% CCTV camera
From Figure \ref{fig:localization-accuracy}b we see a trend similar to the plot in \ref{fig:localization-accuracy}a. The farther away a person is from the origin ($o_{gnd}$), the greater the error between the ground truth and the pedestrian's estimated location. This is due to the approximations in the homography in obtaining the top view from the angled CCTV view, which carries forward to computing [$x^{P_a}_{feet, top}, y^{P_a}_{feet, top}$] (Section \ref{ped-localization-cctv}). However, the maximum error between the estimates and ground truths is again within 0.25 meters. Also, since a pedestrian's location is estimated by the point corresponding to his/her feet, errors due to the pedestrian's orientation are less frequent. The average error in the distance estimation between pedestrians is $\sim$ 0.3 feet.    

\subsubsection{Breach Detection and Enforcement}
 
\begin{table}[]
\centering
\resizebox{\columnwidth}{!}{
\begin{tabular}{|c|c|c|c|}
\hline
\rowcolor{lightgray} \multicolumn{4}{|c|}{Case 1: Static Robot No Occlusions} \\
\hline
\textbf{Metric} & \textbf{CCTV-only} & \textbf{Robot-only} & \textbf{Robot-CCTV Hybrid} \\
\hline
Number of breaches detected & 20 & 10 & 30 \\
% \rowcolor{gray}
Number of enforcements & NA & 10 & 30 \\
\hline
\rowcolor{lightgray} \multicolumn{4}{|c|}{Case 2: Static Robot With 50 \% Occlusion} \\
\hline
% Metric & CCTV-only & Robot-only & CCTV + Robot \\
% \hline
Number of breaches detected & 20 & 7 & 27 \\
% \rowcolor{gray}
Number of enforcements & NA & 7 & 27 \\
\hline
\rowcolor{lightgray} \multicolumn{4}{|c|}{Case 3: Lawnmower exploration With 50\% Occlusions} \\
\hline
% Metric & CCTV-only & Robot-only & CCTV + Robot \\
% \hline
Number of breaches detected & 20 & 20 & 40 \\
% \rowcolor{gray}
Number of enforcements & NA & 20 & 40 \\
\hline
\end{tabular}
}
\caption{\small{Comparison of three configurations in terms of detecting breaches in social distancing norms when two pedestrians are static in any one of 40 points in a laboratory setting. We observe that CCTV + robot configuration has the most number of breaches detected even when the robot is static and outside the CCTV's sensing range. When the robot is mobile, following lawnmower waypoints outside of the CCTV's FOV, it can detect a breach in any of the 20 locations that could not be detected by the CCTV camera.}}
\label{tab:detections-enforcements}
\vspace{-10pt}
 \end{table}

In this experiment we compare the performance differences in detecting a social distancing breach and enforcing social distancing guidelines for three configurations: 1. CCTV only, 2. Robot only, and 3. Robot-CCTV hybrid system. 

The detection and enforcement capabilities of these systems in dynamic scenes vary extensively depending on the initial orientation of the robot and the walking speed and walking directions of pedestrians. Therefore, we standardize the experiment by comparing the best performances of the three configurations in terms of their ability to detect crowding and social distancing breaches in static scenes and the number of times the robot attended to those breaches in a laboratory setting. We demonstrate the robot's ability to track mobile pedestrians in the next section. 

For this experiment, we uniformly sample 40 points in our lab, with 20 points within the FOV of the CCTV camera, and 20 points outside it. Each one of those points could be a location for a social distancing breach at any time instant. We evaluate how many of these points are visible to both cameras and the effect of the robot's mobility. The robot is placed in a fixed location outside the sensing region of the CCTV camera for the static case, and in the mobile case, the robot moves along a lawnmower trajectory outside the CCTV's FOV. The social distancing breaches can also be partially occluded. When a breach is 50\% occluded, we mean a scenario where a person blocks another person such that the half of the human body divided by the sagittal plane is visible to the camera.

The results are shown in Table \ref{tab:detections-enforcements}. As can be seen, the CCTV-only configuration is capable of detecting the standard 20 breaches within its sensing region. It can also handle occlusions between pedestrians better and detect breaches due to the CCTV camera's global view of the environment. It should be noted that this system is an improvement over current CCTV systems where a human manually detects excessive crowding and initiates countermeasures. However, there is no scope for enforcing social distancing at the location of the breaches.

The robot-only configuration detects fewer breaches (10 breaches) than the CCTV setup within the RGB-D camera's sensing region when the robot is static (due to its low FOV). Objects occluding the social distancing breaches also adversely affect the number of detections made by the robot. However, when the robot is moving along a lawnmower trajectory outside the CCTV's FOV, the robot detects the social distancing breaches that could be at any of the 20 locations regardless of whether they are occluded or not.

The robot-CCTV hybrid configuration provides the best performance of the three configurations in terms of detecting novel breaches at the most locations when the robot is static. This is because, when the robot is outside the sensing region of the CCTV camera, the hybrid configuration monitors the largest area in the environment. This configuration also provides better tracking capabilities when a pedestrian is walking (see Section \ref{pursue-walking-ped}). We also note that, in static scenarios, the robot attends to 100\% of the breaches that are detected.

\subsubsection{RGB-D Pedestrian Tracking Duration}
% Table 2
\begin{table}[]
\centering
\resizebox{\columnwidth}{!}{
\begin{tabular}{|c|c|}
\hline
\rowcolor{lightgray} \multicolumn{2}{|c|}{Case 1: Maximum Robot Angular Velocity = 0.5 rad/sec} \\
\hline
\footnotesize{\textbf{Pedestrian Velocity (m/sec)}} & \footnotesize{\textbf{Tracking time (sec)}} \\
\hline
0.25 & 20 (\gray{20}) \\
\hline
0.5 & 6.59 (\gray{10})\\
\hline
0.75 & 3.15 (\gray{6.67})\\
\hline
1 & 2.95 (\gray{5})\\
\rowcolor{lightgray} \multicolumn{2}{|c|}{Case 2: Maximum Robot Angular Velocity = 0.75 rad/sec} \\
\hline
% \small{{Pedestrian Velocity (m/sec)}} & \small{{Tracking time (sec)}} \\
% \hline
0.25 & 20 (\gray{20})\\
\hline
0.5 & 10 (\gray{10})\\
\hline
0.75 & 3.91 (\gray{6.67})\\
\hline
1 & 2.93 (\gray{5})\\
\rowcolor{lightgray} \multicolumn{2}{|c|}{Case 3: Maximum Robot Angular Velocity = 1.0 rad/sec} \\
\hline
% \small{{Pedestrian Velocity (m/sec)}} & \small{{Tracking time (sec)}} \\
% \hline
0.25 & 20 (\gray{20})\\
\hline
0.5 & 10 (\gray{10})\\
\hline
0.75 & 6.58 (\gray{6.67})\\
\hline
1 & 2.77 (\gray{5})\\
\hline
\end{tabular}
}
\caption{\small{The duration for which the robot tracks a walking pedestrian for different pedestrian walking speeds and maximum angular velocities of the robot. The pedestrian walks 5 meters in a direction perpendicular to the robot's orientation and it has to rotate and track the walking pedestrian. The ideal time for which a pedestrian should be tracked is given in the bracket beside the actual time. The robot can effectively track a pedestrian walking at up to 0.75 m/sec when its angular velocity is 1 rad/sec.}}
\label{tab:ped-tracking-vel}
\vspace{-10pt}
 \end{table}
 
\begin{figure*}[t]
      \centering
      \includegraphics[width=\textwidth,height=5.25cm]{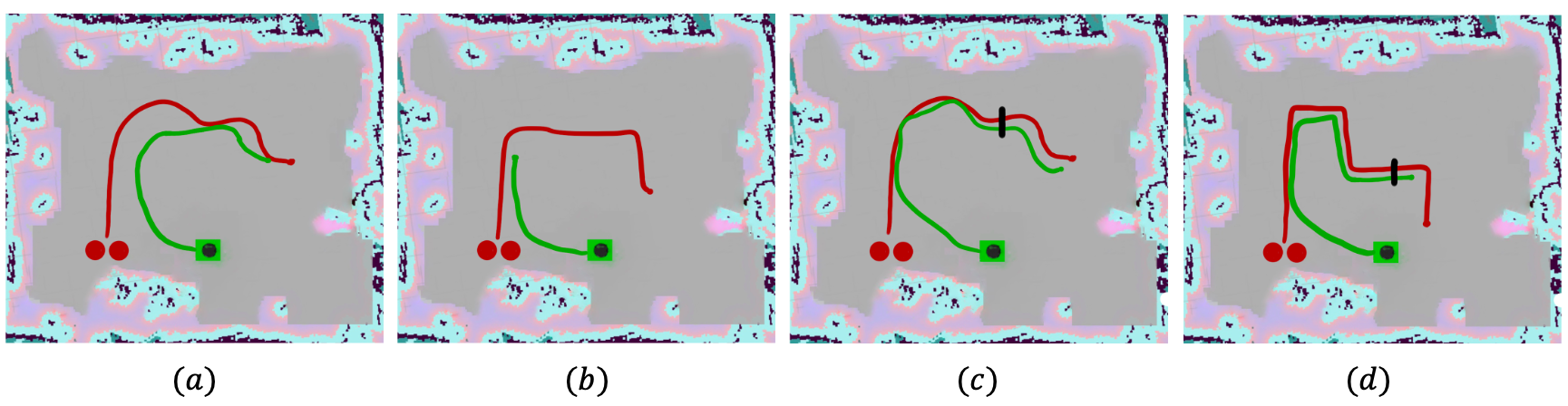}
      \caption {\small{Trajectories of two non-compliant pedestrians (in red) and the robot pursuing them (in green) in the mapped environment shown in figure \ref{fig:3-transformations}c. The pink and blue colors denote the static obstacles in the environment. \textbf{a.} The robot only uses its RGB-D sensors to track the pedestrian. The robot pursues the pedestrians successfully when they move in a smooth trajectory. \textbf{b.} The robot's RGB-D camera is unable to track the pedestrians when they make a sudden sharp turn. \textbf{c.} When the CCTV camera is used to track the pedestrians, the robot follows their trajectories more closely. \textbf{d.} Pedestrians making sharp and sudden turns can also be tracked. The black line denotes the point where the pedestrians leave the CCTV camera's FOV, from where the RGB-D camera tracks the pedestrians. Sharp turns in \textbf{d} again become a challenge.}}
      \label{fig:walking-ped-traj}
    %   \vspace{-15pt}
\end{figure*}

\begin{figure*}[t]
      \centering
      \includegraphics[width=\textwidth,height=4.10cm]{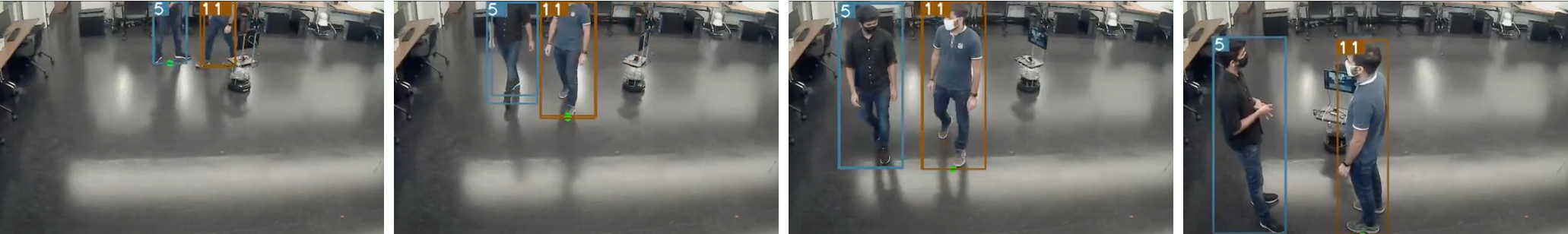}
      \caption {\small{Two mobile non-compliant pedestrians detected by the CCTV camera, pursued by our COVID-Robot in a laboratory setting. The locked pedestrian is marked with a green dot at his feet. Note that the locked pedestrian is changed based on the positions of the two pedestrians in the CCTV footage. The robot pursues them until they maintain the appropriate distance.}}
      \label{fig:ped-pursuit}
    %   \vspace{-15pt}
\end{figure*}
 
In this experiment, we measure the duration for which the robot-only configuration can track walking pedestrians using only its onboard RGB-D sensor. Since it is limited by its FOV, continuously tracking a pedestrian who is walking out of the RGB-D camera's FOV is challenging. To counteract this limitation and track a pedestrian for a longer time, the robot has to rotate/move towards the pedestrian along the pedestrian's walking direction. We vary the walking speed of a pedestrian moving in a direction that is perpendicular to the orientation of the robot. We also vary the maximum angular velocity of the robot to measure the differences in tracking performance (Table \ref{tab:ped-tracking-vel}). 

We observe that the greater the maximum angular velocity of the robot, the better it can track a fast-moving pedestrian. However, since the robot is navigating among humans, we limit the maximum linear and angular velocities to 0.75 m/sec and 0.75 rad/sec, respectively, to minimize the disturbance it causes them. We observe that capping the angular velocity makes it challenging for the robot to track pedestrians walking at $>0.75$ m/sec. Even when the robot is used at its maximum 1 rad/sec angular velocity, pedestrians walking at 1 m/sec are difficult to track. This can only be alleviated in the future when depth cameras improve their range and FOV.  

\subsubsection{CCTV-Guided Walking Locked Pedestrian Pursuit} \label{pursue-walking-ped}
We qualitatively demonstrate how a robot pursues two walking non-compliant pedestrians by plotting their trajectories in the cases where the RGB-D (see figure.\ref{fig:walking-ped-traj}a and b) or the CCTV camera (see figure.\ref{fig:walking-ped-traj}c and d) detects him/her. Figure \ref{fig:walking-ped-traj} a shows that when the pedestrians walk in a smooth trajectory without sharp turns, the robot is able to successfully track them throughout their walk.  

In figure \ref{fig:walking-ped-traj}b, we observe that when the pedestrians make a sharp turn and manage to go outside the limited FOV of the RGB-D camera, the robot is unable to pursue him/her. The pedestrians were walking at speeds $\sim$0.75 m/sec. This issue is alleviated when the CCTV camera tracks both the pedestrians instead of the RGB-D camera. 

Figure \ref{fig:walking-ped-traj}c and d show that the robot is able to track the pedestrians more closely and accurately with the goal data computed using the CCTV's localization. In addition, sudden and sharp turns by the pedestrians are handled with ease, and pedestrians moving at speeds $\sim 0.75$ m/sec can be tracked and pursued, which was not possible with the robot-only configuration. When the pedestrians move out of the CCTV camera's FOV (black line in figures \ref{fig:walking-ped-traj}c and d), the data from the robot's RGB-D camera helps pursue the two pedestrians immediately. However, the pedestrians' sharp turns again becomes a challenge to track. 

The robot pursuing two non-compliant pedestrians in our lab setting is shown in figure \ref{fig:ped-pursuit}.

\section{Conclusions, Limitations and Future Work}
We present a novel method to detect breaches in social distancing norms in indoor scenes using visual sensors such as RGB-D and CCTV cameras. We use a mobile robot to attend to the individuals who are non-compliant with the social distancing norm and to encourage them to move apart by displaying a message on a screen mounted on the robot. We demonstrate our method's effectiveness in localizing pedestrians, detecting breaches, and pursuing walking pedestrians. We conclude that the CCTV+robot hybrid configuration outperforms configurations in which only one of the two components is used for tracking and pursuing non-compliant pedestrians. % Add about thermal camera and 

Our method has a few limitations. For instance, our method does not distinguish between strangers and people from the same household. Therefore, all individuals in an indoor environment are encouraged to maintain a 6-foot distance from each other.  Our current approach for issuing a warning to violating pedestrians using a monitor has limitations, and we need to develop better human-robot approaches. As more such monitoring robots are used to check for social distances or collecting related data, this could also affect the behavior of pedestrians in different settings. 

We need to perform more studies on the social impact of such robots. Due to COVID restrictions, we have only been able to evaluate the performance of COVID-robot in our low to medium density laboratory settings. Eventually, we want to evaluate the robot's performance in crowded public settings and outdoor scenarios. We also need to design better techniques to improve the enforcement of social distancing by using better human-robot interaction methods.

\section*{Acknowledgment}
This work is supported in part by ARO grant W911NF1910315 and NSF grant 2031901.

\ifCLASSOPTIONcaptionsoff
  \newpage
\fi

\bibliographystyle{IEEEtran}
\bibliography{References}
\end{document}